\documentclass{article}

\usepackage{microtype}
\usepackage{graphicx}
\usepackage{subfigure}
\usepackage{stfloats}
\usepackage{comment}
\usepackage{mathtools}
\usepackage{multirow}   
\usepackage{amsmath,amssymb,amsfonts}
\usepackage{booktabs} % for professional tables
\usepackage{hyperref}

\usepackage[accepted]{icml2021}

\icmltitlerunning{TD-GEM}
\usepackage[printonlyused]{acronym}
       \acrodef{GAN}{Generative Adversarial Network}
    \acrodef{VTON}{Virtual Try-ON}
    \acrodef{CLIP}{Contrastive Language-Image Pre-training}
    \acrodef{SHHQ}{Stylish-Humans-HQ}
    \acrodef{TD-GEM}{Text-Driven Garment Editing Mapper}
    \acrodef{e4e}{Encoder for Editing}
    \acrodef{PTI}{Pivotal Tuning Inversion}
    \acrodef{FFCLIP}{Free-Form CLIP}
    \acrodef{VQVAE}{Vector-Quantized Variational Autoencoder}
    \acrodef{FID}{Fréchet Inception Distance}
    \acrodef{SSIM}{Structural Similarity Index Measure}
    \acrodef{PSNR}{Peak Signal-to-Noise Ratio}
    \acrodef{ACD}{Average Color Difference}

\begin{document}

\twocolumn[
\icmltitle{TD-GEM: Text-Driven Garment Editing Mapper}



\icmlsetsymbol{equal}{*}

\begin{icmlauthorlist}
\icmlauthor{Reza Dadfar}{equal}
\icmlauthor{Sanaz Sabzevari}{equal}
\icmlauthor{Mårten Björkman}{}
\icmlauthor{Danica Kragic}{}
\\
  School of Electrical Engineering and Computer Science, KTH Royal Institute of Technology\\
Stockholm, Sweden\\
  \texttt{\{dadfar,sanazsab,celle,dani\}@kth.se}\\
\end{icmlauthorlist}

%\icmlaffiliation{to}{School of Electrical Engineering and Computer Science, KTH Royal Institute of Technology, Stockholm, Sweden}

%\icmlcorrespondingauthor{Sanaz Sabzevari}{sanazsab@kth.se}
%\icmlcorrespondingauthor{Reza Dadfar}{dadfar@kth.se}

\icmlkeywords{Machine Learning, ICML}

\vskip 0.3in
]



\printAffiliationsAndNotice{\icmlEqualContribution}  % leave blank if no need to mention equal contribution

\begin{abstract}
Language-based fashion image editing allows users to try out variations of desired garments through provided text prompts.
%on their desired garment with one text prompt digitally. 
Inspired by research on manipulating latent representations in StyleCLIP and HairCLIP, we focus on these latent spaces for editing fashion items of full-body human datasets. Currently, there is a gap in handling fashion image editing due to the complexity of garment shapes and textures and the diversity of human poses. In this paper, we propose an editing optimizer scheme method called~\ac{TD-GEM}, aiming to edit fashion items in a disentangled way. To this end, we initially obtain a latent representation of an image through generative adversarial network inversions such as ~\ac{e4e} or ~\ac{PTI} for more accurate results. An optimization-based~\ac{CLIP} is then utilized to guide the latent representation of a fashion image in the direction of a target attribute expressed in terms of a text prompt. Our TD-GEM manipulates the image accurately according to the target attribute, while other parts of the image are kept untouched. In the experiments, we evaluate~\ac{TD-GEM} on two different attributes (\textit{i.e.,} “color" and “sleeve length"), which effectively generates realistic images compared to the recent manipulation schemes.
\end{abstract}

%\begin{figure}[ht]
%\vskip 0.2in
%\begin{center}
%\centerline{\includegraphics[width=\columnwidth]{icml_numpapers}}
%\caption{Historical locations and number of accepted papers for International
%Machine Learning Conferences (ICML 1993 -- ICML 2008) and International
%Workshops on Machine Learning (ML 1988 -- ML 1992). At the time this figure was
%produced, the number of accepted papers for ICML 2008 was unknown and instead
%estimated.}
%\label{icml-historical}
%\end{center}
%\vskip -0.2in
%\end{figure}

\section{Introduction}
Text-driven garment editing frameworks provide a convenient digital tool for end-users to edit fashion items. %The synthesized image quality and user-friendliness %significantly improve the sustainable online
The application of high-quality synthesized images for visualization of not yet produced garments allows for a more sustainable online
fashion industry, ultimately decreasing the retailer's costs and environmental carbon footprint \cite{kozlowski2012environmental}. Recently,~\acp{GAN} \cite{goodfellow2014generative} have been used for generating photo-realistic images for various datasets. It is extensively employed in~\acp{VTON} \cite{wang2018toward,dong2019towards,neuberger2020image,visapp23} and outfit generators \cite{jiang2022text2human}. Despite tremendous development in this domain, text-conditioned human outfit editing has not yet been well explored.

Image attribute manipulation  requires an accurate latent mapping between the text embedding space and latent visual space of the synthesized image often implemented by StyleGAN-based \sloppy approaches \cite{karras2019style,karras2020training,karras2020analyzing,roich2022pivotal}. Recently, this has been done in pioneering studies such as StyleCLIP \cite{patashnik2021styleclip} and TediGAN \cite{xia2021tedigan} to edit images based on a target text prompt. They find the latent visual subspace aligned to the text embedding space. However, StyleCLIP attains text-based semantic image editing through Contrastive Language-Image Pre-training~(CLIP)\acused{CLIP} encoding \cite{radford2021learning}. To generate high-quality images, StyleGAN2 \cite{wang2018toward} has shown great promise across various applications. The majority of research studies have focused on face, car, and building datasets despite limiting exploration in the domain of human clothing \cite{fu2022stylegan}. Due to the diverse range of human poses and intricate textures and shapes of garments, manipulating human outfits through generative models is a challenging task \cite{lee2020maskgan}. Thanks to work by \cite{fu2022stylegan}, a large-scale fashion image dataset called~\ac{SHHQ} was collected and trained through the StyleGAN2 network. 

This paper addresses image manipulation, including text-conditioned editing in the fashion domain using the~\ac{SHHQ} dataset. We learn a mapping between text prompt embeddings and latent representations of input images while generating disentangled output images based on text descriptions. The proposed Text-Driven Garment Editing Mapper~(TD-GEM)\acused{TD-GEM} edits attributes of the input image according to the input text, \textit{e.g.} the sleeve length or color of the garment, using a single mapper and inversion space. It successfully preserves the irrelevant attributes of the input image. The primary contributions of this paper are

\begin{itemize}
\item We provide a text-driven image manipulation framework, TD-GEM, for full-body fashion images using~\ac{CLIP} and~\ac{GAN} inversion. 

\item We improve the speed of the process by training a single network for each input text rather than solving an optimization problem per image.

\item \ac{TD-GEM} consists of a modulation network, acting in a disentangled semantic space, that allows changes in \textit{e.g.,} color and sleeve length based on user requests. 

\end{itemize}

\section{Related work}
\subsection{Text-Conditioned Image Editing}
There is a vast array of studies on image editing in the literature \cite{ling2021editgan} for various datasets, but we focus on using text prompts as input in image manipulation here, like \cite{reed2016generative}. Starting from \cite{dong2017semantic} that employs natural language description to synthesize images conditioned on the given text and image embeddings, several works \cite{nam2018text, shen2020interpreting} enhance the quality of synthesized images and disentangle visual attributes. To make it more concrete, \cite{nam2018text} proposes text-adaptive~\ac{GAN}, creating world-level local discriminators based on the text prompt to represent a visual attribute. It leads to generating images that only modify regions associated with the given text. Another line of work \cite{shen2020interpreting} focuses on semantic editing using a novel~\ac{GAN} called ManiGAN to preserve irrelevant content in the input image. It entails two modules to initially select regions relevant to the input text and then refine missing contents of the synthetic image. It is worth noting that ManiGAN only applies to the CUB and COCO datasets. InterFaceGAN \cite{shen2020interfacegan} interprets a~\ac{GAN} model for disentangled and controllable face representation and identifies facial semantics encoded in the latent space. Although a simple, effective approach for face editing, it aligns attributes with a linear subspace of the latent space resulting in failure for long-distance manipulation. 

An alternative perspective on this area of research is to manipulate the image using visual-semantic alignment or image-text matching rather than word-level training feedback. In this approach, semantics are mapped from text to images. Aligned to this strategy, TediGAN \cite{xia2021tedigan} generates diverse and high-quality images using a control structure based on style-mixing with multi-modal inputs such as sketches or text prompts. It can manipulate images with particular attributes through the common latent space of input text and images.
Another improved approach to discovering semantically latent manipulation without using an annotated collection of images is explored by StyleCLIP \cite{patashnik2021styleclip}. It
develops a text-guided latent manipulation for StyleGAN image manipulation, using~\ac{CLIP} in an optimization scheme as a loss network. Benefiting from image text representation like StyleCLIP, human hair editing is introduced by \cite{wei2022hairclip} referred to as HairCLIP. It trains a mapper network to map the input references into embedded latent code and exploits the text encoder and image decoder of~\ac{CLIP}. Recently, the latent mapping between the StyleGAN latent space and the text embedding space of CLIP is designed in \cite{zhu2022one}, introducing~\ac{FFCLIP} to handle free-form text prompts. It leverages input text with multiple semantic meanings to edit images. Nevertheless, it is worth noting that certain disentanglement challenges remain due to the presence of human biases \cite{fu2019dual}. 

\subsection{Image Synthesis and Editing in Fashion domain}
Manipulating images in the fashion industry becomes increasingly complicated when dealing with the full human body, compared to editing images of specific body parts, like the face or hair. Typically, an image editing pipeline involves translating an input image into a latent space representation using inversion techniques, followed by decoding the modified latent representation to generate an output image \cite{fu2022stylegan}. In the context of the fashion domain, another essential aspect to consider is establishing a proper mapping from the garment to the human body while preserving the identity of humans and the rest of the fashion items in the original images. The pioneering work of   
\cite{zhu2017your} known as FashionGAN approaches end-to-end virtual garment display by training a conditional~\ac{GAN} \cite{mirza2014conditional}. FashionGAN trains an encoder using the real fabric pattern image, which results in the latent vector containing solely the material and color information of the fabric pattern image. Afterwards, a supplementary local loss module is integrated to regulate the texture synthesis process carried out by the generator. Another concurrent work to FashionGAN, \cite{rostamzadeh2018fashion}, provides a rich dataset, including extensive annotations called Fashion-Gen applied for the text-to-image task. However, the quality of synthesized images using the StackGAN method is blurry, especially for a face. Another aforementioned dataset (\ac{SHHQ}) collected by \cite{fu2022stylegan} is utilized for StyleGAN-based structures. Their investigation focused on analyzing how various factors, such as the size of the data set, the distribution of the data, and the alignment of the person in the image, influence the quality of generated images. To validate editing techniques with this dataset, SOTA facial StyleGAN-based architectures like InterFaceGAN \cite{shen2020interfacegan}, StyleSpace \cite{wu2021stylespace}, and SeFa \cite{shen2021closed} are evaluated. Text2Human \cite{jiang2022text2human} also uses the~\ac{SHHQ} dataset and generates synthesized images by applying a human posture, a textual description of the garment's texture and shape as inputs. To encode the images, they implemented a hierarchical~\ac{VQVAE} framework \hspace{1sp}\cite{esser2021imagebart} with a texture-aware codebook. In contrast to earlier research, which constrains the verbal proficiency of the input text owing to sparsing the text into a closed set of categories, a recent work named FICE~\cite{pernuvs2023fice} suggests a latent-code regularization approach using a text-conditioned editing model. While the FICE model performs high-quality image editing, experiments on full-body human images are not explored.

\section{Fashion Image Editing}

To achieve image manipulation, it is necessary to obtain a latent representation of the source image within the latent space, which can be carried out by~\ac{GAN} inversion. Then images can be edited using methodologies such as latent optimizer, StyleCLIP mapper, or our proposed mapper network. Our approach involves a two-stage process for image editing, wherein we explore the surrounding area of the latent code to identify a latent representation corresponding to the edited image through a loss function. Once this representation has been determined, we input the code into the~\ac{GAN} architecture to generate the desired edited image.

\subsection{GAN Inversion}  
A crucial aspect of an effective inversion approach is its ability to balance the trade-off between distortion and editability. Specifically, the method should be capable of preserving the original appearance of an image (\textit{i.e.,} low distortion) while enabling convincing attribute modifications (\textit{i.e.,} high editability). One such method that claims to achieve this goal is~\ac{PTI}, as introduced by \cite{roich2022pivotal}. This technique employs an off-the-shelf encoder, such as~\ac{e4e}, to derive a latent code for the StyleGAN architecture. However, the encoder's output can result in distortion in the reconstructed image compared to the original, which is known as the identity gap. To address this issue, they fine-tuned the generator to preserve the image's identity. The process of adjustment can be analogized to the act of aiming a dart towards a target and subsequently realigning the board to account for a near-miss.

In this paper, we utilized a pre-trained~\ac{e4e} encoder for the inversion, as provided by \cite{fu2022stylegan}, with further fine-tuning of the generator based on a specific loss term. The loss function $\mathcal{L}(\theta)$ was defined as the sum of the learned perceptual image patch similarity loss function ($\mathcal{L}_{LPIPS}$) and the pixel-wise mean square error ($\mathcal{L}_2$),
\begin{multline}
    \mathcal{L}(\theta)=\frac{1}{N}\sum_{i=1}^{N}
    \mathcal{L}_{LPIPS}(x_i, G(w_i:\theta)) \\
   + \lambda_{2}\mathcal{L}_2(x_i, G(w_i:\theta)),
\end{multline}
with a hyperparameter $\lambda_2$ set to $1$. The aim of the optimization is to determine the optimal parameters $\theta^*$ for the generator $G$, based on the output of the~\ac{e4e} encoder $w_i$ for each image $x_i$ in the dataset of size $N$. The fine-tuned generator $G_{\theta^*}$ is henceforth denoted as $G$ for enhancing the readability. We used the AlexNet network to calculate the perceptual loss, with the learning rate set to $5\times10^{-4}$, a maximum number of iterations, $3500$, and a convergence tolerance of $10^{-4}$.

\subsection{Latent Optimizer}
\label{sec: Latent optimizer}
\begin{figure}[b]
  \begin{center}
    \includegraphics[width=80mm]{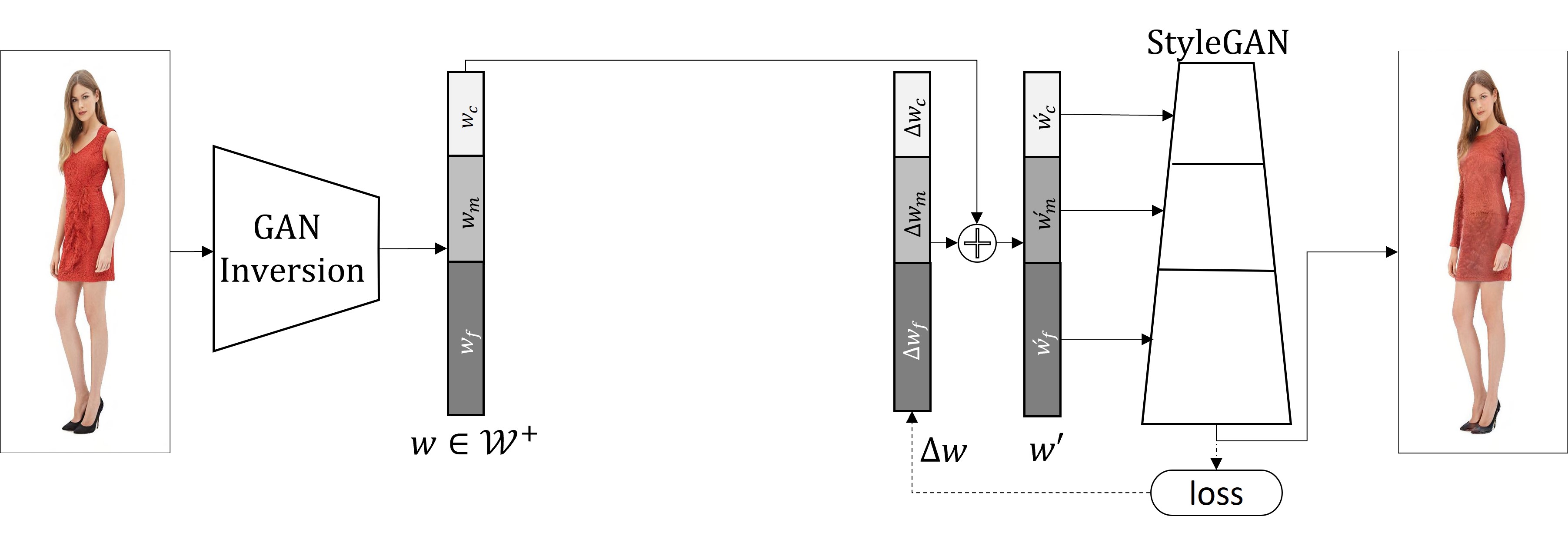}
  \end{center}
  \caption[I]{The manipulation of garments through lengthening sleeves via the use of the latent optimizer approach.}
  \label{fig:op}
\end{figure}

The latent optimizer framework is an image manipulation approach that relies solely on solving a direct optimization problem \cite{patashnik2021styleclip}. This framework uses the GAN inversion to first invert an image into a latent code. Then, an optimization problem is solved using a loss function to find the latent code residual. The residual is added to the original latent representation and fed into the StyleGAN to obtain the edited image (see Figure \ref{fig:op}).

The loss function is as follows:
\begin{align}
    \begin{split}
        \mathcal{L}_{t} =& \lambda_\text{CLIP} \mathcal{L}_\text{CLIP} + \lambda_{2} \|\Delta w \|_2+\lambda_{ID} \mathcal{L}_{ID} 
         \label{eq:lossLatent}
    \end{split}
\end{align} 

The clip loss $\mathcal{L}_\text{CLIP} $ is designed to guide the optimization process toward achieving the attribute described in the input text. To accomplish this, the embeddings of both the input text ($t$) and the generated image ($G(w')$) are obtained in a shared space using the pre-trained CLIP encoder, and their cosine similarity is considered in the loss function:
\begin{align}
    \mathcal{L}_\text{CLIP}  = 1 - cos(E_\text{CLIP} (G(w^{\prime})), E_\text{CLIP} (t)),
\label{eq:LossCLIP}
\end{align}

The term $\|\Delta w \|_2$ ensures exploration within the vicinity of the original latent representation. This term guarantees a localized manipulation of the initial image.

To preserve the identity of the original image, an identity loss term $\mathcal{L}_{ID}$ is used. It calculates the mean square error between the features of the original and edited images obtained from the last layer of a pre-trained ConvNeXt network.
\begin{align}
    \mathcal{L}_{ID} = MSE(R(G(w)), R(G(w^{\prime}))). 
    \label{eq:LossID}
\end{align}
where $MSE$ is the mean square error, $R$ is the pre-trained ConvNeXt network~\cite{liu2022convnet}, and $R(G(w))$ and $R(G(w^{\prime}))$ are the features of the ground truth and generated images, respectively.

\subsection{StyleCLIP Mapper}
The latent optimization framework is not efficient in terms of image editing due to the need to solve an optimization problem for each image. To address this issue,~\cite{patashnik2021styleclip} introduced a mapper network that can infer the manipulated image based on a given input text, making the process more efficient.
%==================================
%              Figure
%==================================
\begin{figure}[b]
  \begin{center}
    \includegraphics[width=80mm]{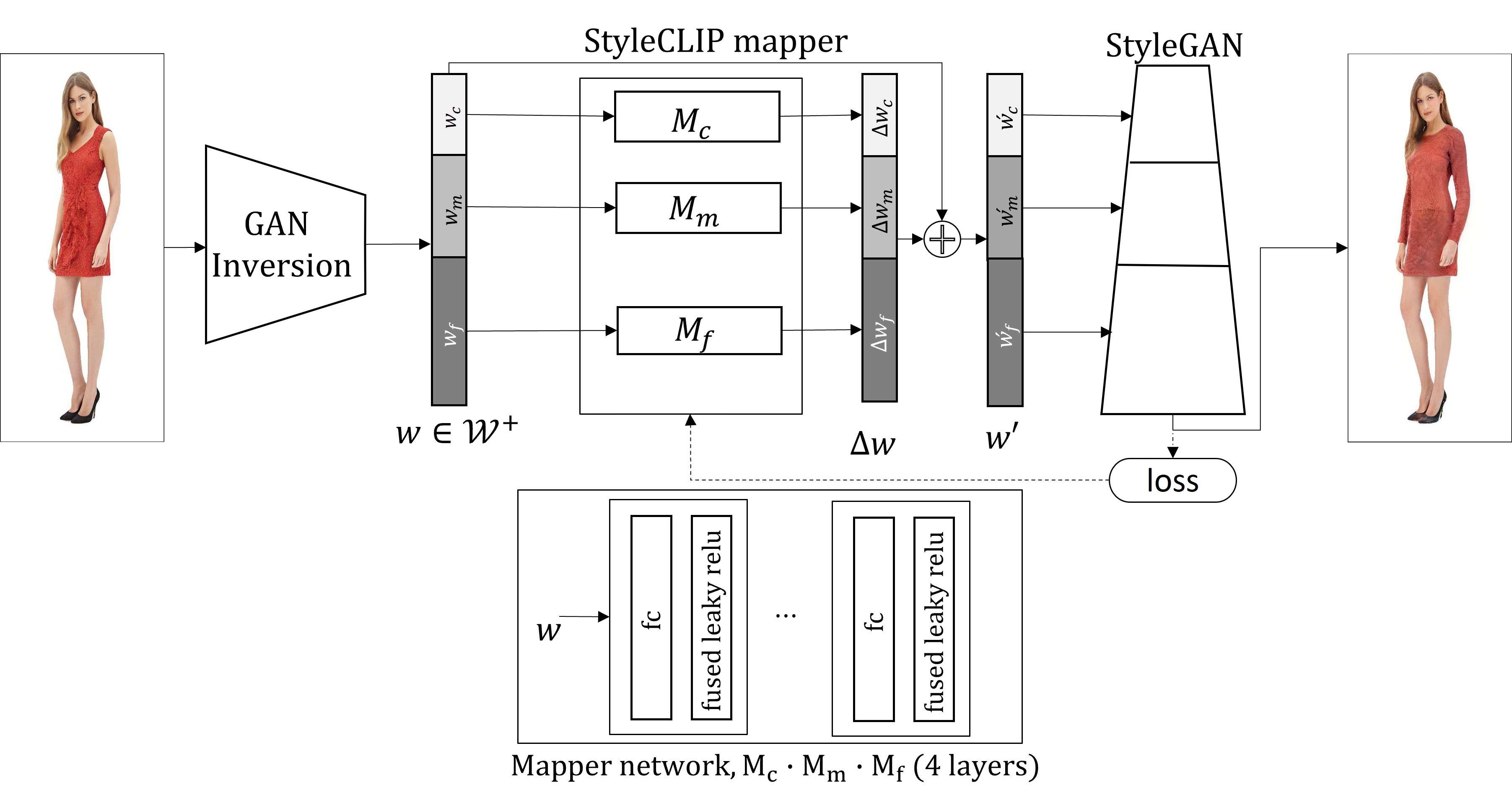}
  \end{center}
  \caption[I)]{The StyleCLIP mapper network is utilized to increase the length of sleeves.}
  \label{fig:arch_mapper}
\end{figure}
%==================================
The mapper network is first trained on the training dataset. It is then used to edit new images from the testing dataset. The architecture of the mapper is designed with three distinct sub-modules, each responsible for different aspects of the generated image (Figure \ref{fig:arch_mapper}). These sub-modules are divided into coarse, medium, and fine clusters, which control the corresponding structures in the image. A more detailed description of the architecture can be found in~\cite{patashnik2021styleclip}. The input image is first inverted during editing, and the resulting latent code is fed into the mapper to obtain a residual latent code. This residual latent code is then added to the original latent code and passed through the pre-trained StyleGAN generator to produce the edited image. The loss function is defined as:
\begin{align}
    \begin{split}
        \mathcal{L}_{t} =& \lambda_\text{CLIP} \mathcal{L}_\text{CLIP} + \lambda_{2} \|M_t(w) \|_2+\lambda_{ID} \mathcal{L}_{ID} 
         \label{eq:lossStyleCLIP}
    \end{split}
\end{align} 
where $M_t(w)$ is the output of the mapper designed for the input text, $t$. The clip and identity losses are the same as described in section \ref{sec: Latent optimizer}.

\subsection{TD-GEM}
In this paper, we introduce~\ac{TD-GEM} as an innovative approach to manipulating garment attributes using a single mapper network. We drew inspiration from the HairCLIP technique developed by~\cite{wei2022hairclip}, which enhanced previous works such as StyleCLIP. HairCLIP uses additional loss functions and changes the mapper architecture to enable accurate attribute manipulation while preserving irrelevant parts of the image. 

%==================================
%              Figure
%==================================
\begin{figure*}[t]
\centering
\includegraphics[width=\textwidth]{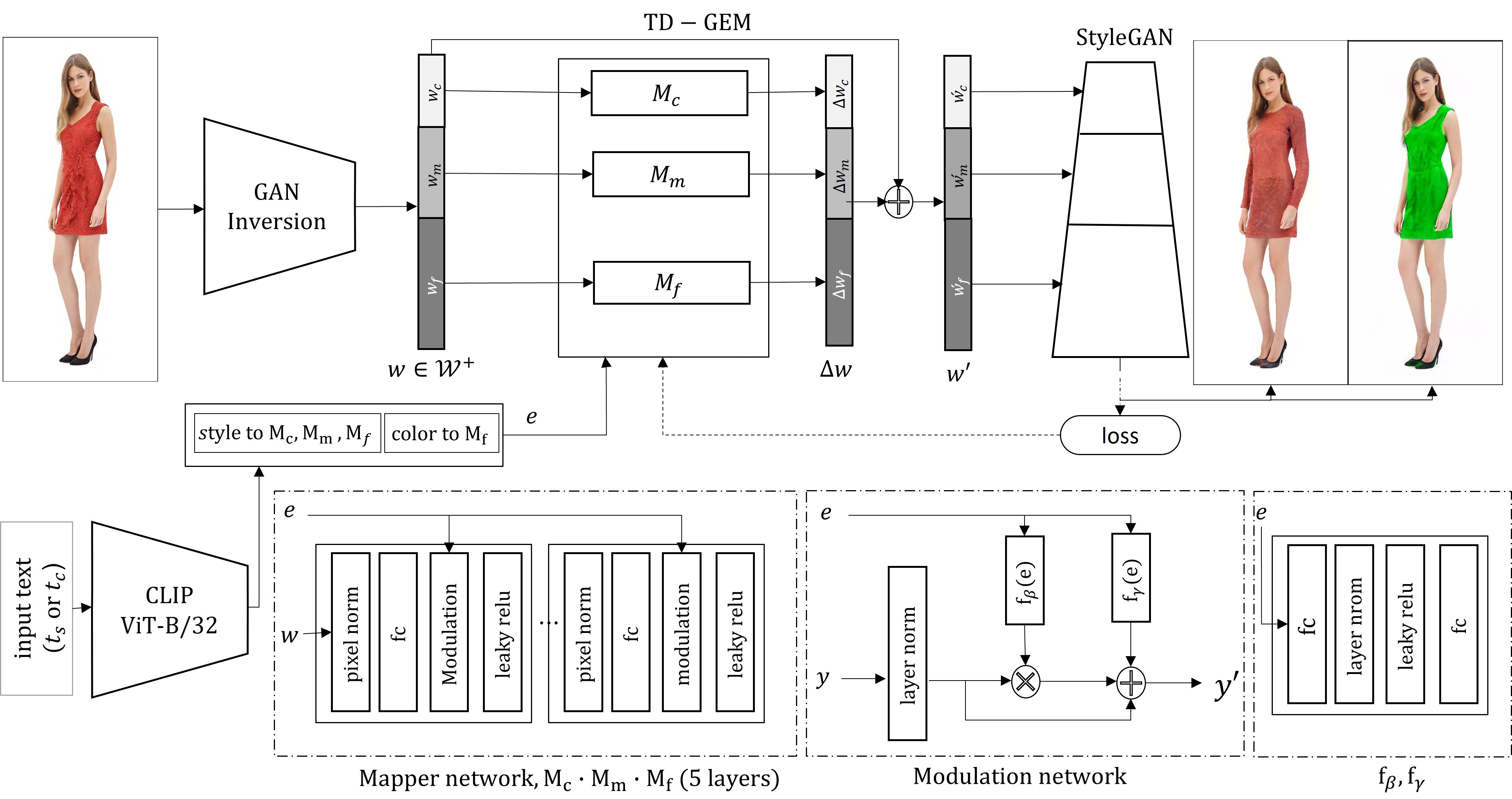}
\vspace*{-3mm}
\caption[TD-GEM framework]{The~\ac{TD-GEM} architecture includes an input image that undergoes inversion via PTI. The output is then passed through the mapper network to obtain a residual,$\Delta w$. The mapper network is composed of three parts, all of which receive text-conditioned input related to the clothing's form and shape. The final part also receives color-conditioned information. The resulting latent code, $w^{\prime}$, is subsequently fed into a pre-trained StyleGAN generator to produce the edited image. The loss function is designed to modify the image's attributes as described in the text while preserving the irrelevant parts.}
\label{fig:arch_hairclip}
\end{figure*}
%\vfill
%==================================

TD-GEM extends HairCLIP's capabilities by adapting the loss functions to the fashion domain allowing for simultaneous editing of both the length of the sleeves and the color of the clothing using a single mapper network. The mapper's definition is expressed as $M = (M_c, M_m, M_f)$. The structure of the mapper is comprised of three distinct sub-modules, each represented as $M_c$, $M_m$, and $M_f$. These sub-modules correspond to the varying degrees of detail present in the images that are generated (Figure \ref{fig:arch_hairclip}). We inject information about the form and shape of the clothing ($t_s$) into all three sub-modules and color information ($t_c$) into the final one. This is in contrast to HairCLIP, where hairstyle information is injected into the first two layers and color information into the last one.
Our mapper receives the latent code of an image through a~\ac{GAN} inversion operation, such as~\ac{PTI}, and text embeddings obtained by feeding the textual description into the encoder of a pre-trained CLIP network.  %We then provide this information to the coarse, medium, and fine sub-modules.
%The text embeddings are injected into the three sub-modules,  while color information is only injected into the fine module ($M_f$). 
Based on this methodology, the mapper produces a residual latent code $\Delta w$. This code is then added to the original latent code of the image, resulting in $w'=w+\Delta w$, and provided to the StyleGAN architecture to generate the edited image. 
%The mapper can be written as:
%$$
%    M(w, E_{clip}(t_s), E_{clip}(t_c)) = (M_c(w_c, E_{clip}(t_s)), %\\ M_m(w_m, E_{clip}(t_s)), M_f(w_f, E_{clip}(t_c)) 
%$$
Each sub-mapper consists of five layers, each consisting of a pixel-norm, fully connected layer, modulation layer, and leaky ReLU activation. The modulation layer of the network encodes the text input information and receives the information from text embeddings and  the previous fully connected layer~\cite{wei2022hairclip}. It processes the text input:
\begin{align}
    y^{\prime} = 1 + f_{\gamma}(e)\frac{y - \mu_y}{\sigma_y} + f_{\beta}(e) \label{eq:Reluoutput}
\end{align}
where  $y^{\prime}$ in the output of the modulation network, $e$ refers to the text embedding obtained from the input text $t$, the parameters $\mu_y$ and $\sigma_y$ represent the mean and standard deviation of the intermediate feature $y$, $f_{\gamma}$ and $f_{\beta}$ are neural networks that consist of a fully connected layer, a layer normalization, a leaky ReLU activation function, and another fully connected layer.

The modulation layer provides semantic alignment. For example,
when the network is given two colors, such as blue and green, the modulation layers respond differently to each color. The architecture is illustrated in Figure \ref{fig:arch_hairclip}.

To enhance manipulation accuracy and encourage disentanglement during the editing process, the following loss function is utilized for training the \ac{TD-GEM} mapper network. 
\begin{align}
    \begin{split}
        \mathcal{L}_{t} =& \lambda_\text{CLIP} \mathcal{L}_\text{CLIP} + \lambda_{2} \mathcal{L}_{norm}+\lambda_{ID} \mathcal{L}_{ID}\\
         &+ \lambda_{color}\mathcal{L}_{color} + \lambda_{BG}\mathcal{L}_{BG} 
         \label{eq:lossTD_GEM}
    \end{split}
\end{align} 
The clip and identity loss functions are as prescribed before. 
The locality of the edit is ensured by $\mathcal{L}_{norm}$ loss as
\begin{align}
    \mathcal{L}_{norm} = \|M(w, E_\text{CLIP} (t))\|_{2}, 
    \label{eq:Lossnorm}
\end{align}
The color composition of the clothing is maintained by introducing a color loss $\mathcal{L}_{color}$ in the loss function. A pre-trained parsing network $P$ \cite{BowenParsing2021} is employed to segment the human instance into foreground and background parts. The foreground area of the image comprises the shirt, dress, coats, neck, and arms, and the rest is assumed as the background. The color loss is only applied to the foreground as
\begin{align}
    \begin{split}
    \mathcal{L}_{color}=&\|avg(G(w^{\prime})\cdot P(G(w^{\prime}))) \\
     &-avg(G(w)\cdot P(G(w)))\|_1, 
    \label{eq:Losscolor}
    \end{split}
\end{align}
where $\|\cdot\|_1$ is the $L_1$ norm $G(w)$ is the original image, $G(w^\prime)$ is the edited image, $G() \cdot P()$ is the foreground and $avg$ is the mean value for each channel. The image is first transformed from $RGB$ color to $XYZ$ and then to $LAB$ coordinates to obtain the average.

To ensure that irrelevant parts of the image are kept untouched, a background loss is applied. It is defined as
\begin{equation}
   \mathcal{L}_{BG}=\|(G(w^{\prime}) - G(w)) \cdot (\neg P(G(w^{\prime}))\cap \neg P(G(w)))\|_2.
   \label{eq:Lossbackground}
\end{equation}
where $\|\cdot\|_2$ denotes the $L_2$ norm. During training, we obtain foreground masks for both the original and edited images at each iteration. We combine these masks to obtain a union, which represents the editable area, while the rest of the image constitutes the background. Mathematically, we express the background as $\neg (P(G(w^{\prime}))\cup P(G(w)))$, which is equivalent to $\neg P(G(w^{\prime}))\cap \neg P(G(w)))$. By applying the $L_2$ norm in the background loss, we ensure that the original and edited images are similar for the areas included in the background.

\section{Experiments}
\subsection{Dataset}
We employ~\ac{SHHQ} dataset for the experiments in this paper. It consists of high-quality, full-body fashion, human-centric images. It contains $230K$ fashion images in diverse poses and textures, with a resolution of $1024 \times 512$ pixels. However, only $40K$ images are presently accessible to external researchers. We have selected 2200 samples with mostly short-sleeved or sleeveless attributes to either lengthen sleeves or change the color. The dataset is split into a 90/10 ratio for training and testing, using 2000 samples for training and 200 for testing.

\subsection{Implementation Detail}
For this experiment, the development is carried out within a dockerized environment utilizing a NVIDIA GeForce 3090 GPU with 24GB VRAM. The code is implemented using PyTorch 1.9.1 and Cuda 11.4. In this work, we train a mapper, $M$, to manipulate images, leveraging a pre-trained StyleGAN2-ADA generator $G$~\cite{karras2020analyzing}, a pre-trained CLIP model~\cite{radford2021learning}, and a pre-trained parsing network, $P$~\cite{BowenParsing2021}.

Table \ref{tab:tab_coeff} presents the coefficients for the loss function employed in this study. Notably, these coefficients differ when editing sleeve length and garment color, with a higher background coefficient used during color editing. The training process utilizes 100k steps (\textit{e.g.}, 50 epochs) and a learning rate of $5\times 10^{-4}$ with a “Ranger" optimizer.
%==================================
%              Table
%==================================
\begin{table}[bp!]
\caption[I]{Coefficients corresponding to various terms within the loss function.}
\vspace*{2mm}
\label{tab:tab_coeff}
\centering
\begin{tabular}{l c c c c c}
Case & \multicolumn{1}{c}{$\lambda_\text{CLIP}$} & \multicolumn{1}{c}{$\lambda_{2}$}  & \multicolumn{1}{c}{$\lambda_\text{ID}$} & \multicolumn{1}{c}{$\lambda_\text{color}$}  & \multicolumn{1}{c}{$\lambda_\text{BG}$}\\
\midrule
sleeve & 1.0 & 1.0 & 1.0 & $5\times 10^{-3}$ & 0.3 \\
color & 1.0 & 1.0 & 1.0 & $5\times 10^{-3}$ & 1 \\
\midrule
\end{tabular}
\end{table}
\subsection{Comparisons and Evaluation}
\subsubsection{Pivotal Tuning Inversion (PTI)}
\begin{figure}[b]
  \begin{center}
    \includegraphics[width=80mm]{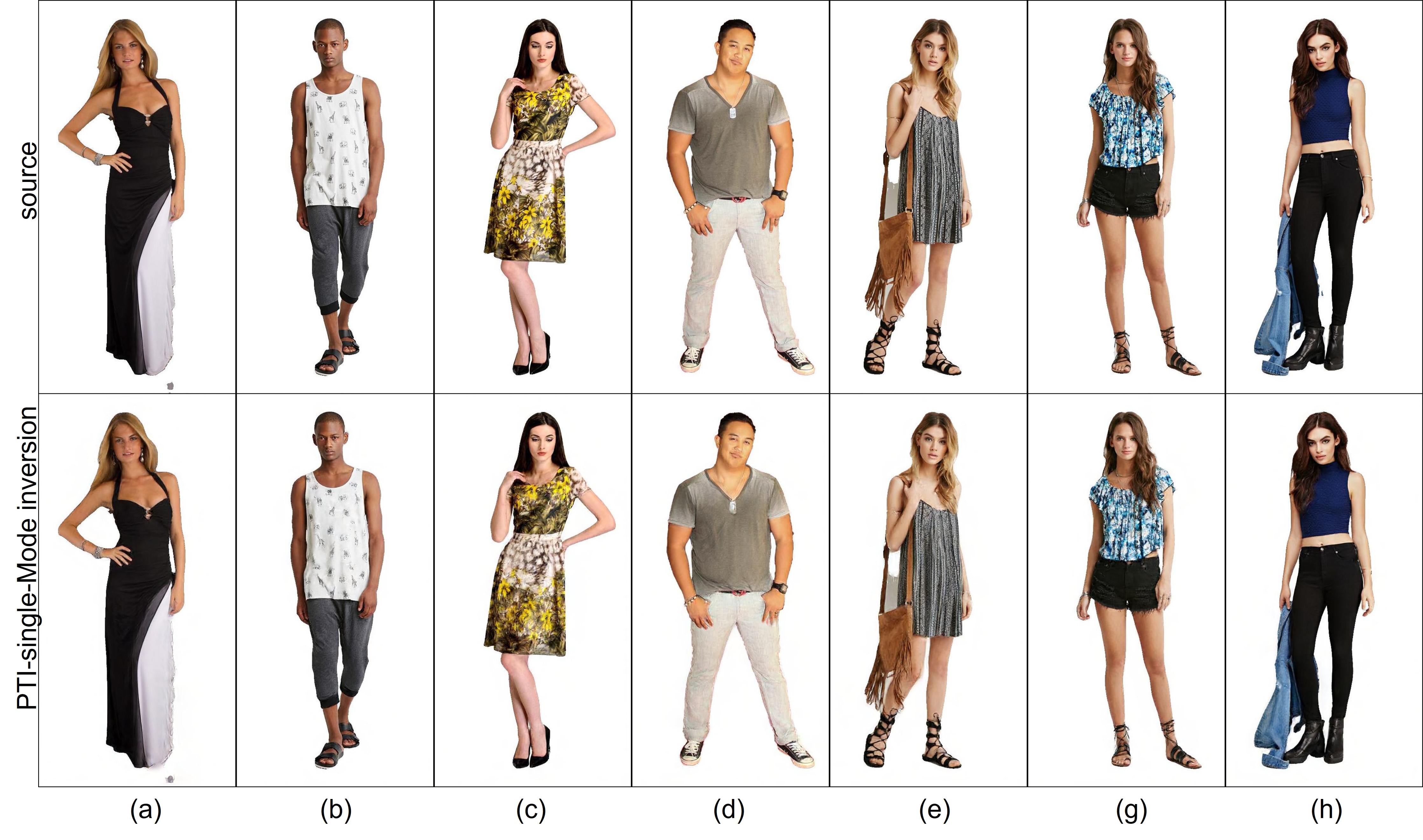}
  \end{center}
  \caption[Images of GAN inversion with PTI single-mode]{This is a collection of qualitative outcomes for PTI single-mode inversion, where the top row displays the original image, and the bottom row exhibits the corresponding inverted ones.} 
  \label{fig:res_inv_pti}
\end{figure}

This section presents the results of PTI, one of the~\ac{GAN} inversion approaches. Figure~\ref{fig:res_inv_pti} displays a selection of sample images obtained using~\ac{PTI}. The method successfully preserves fine-grained details, such as facial features and hair, while maintaining the correct configuration of the human body, including the posture of hands, legs, and shoes. Additionally, the clothing attributes, including the shape, color, and patterns, are accurately preserved, with no missing fashion items in the inverted results. These findings demonstrate the effectiveness of~\ac{PTI} in achieving high-quality GAN inversion with a faithful representation of the input images' attributes.
%==================================
%              Figure
%==================================
%==================================
The quantitative analysis further supports the superiority of the PTI approach over~\ac{e4e}, as shown in Table~\ref{tab:tab_inv_pti}. 
%==================================
%              Table
%==================================
\begin{table}[htb!]
\caption[Quantitative results of GAN inversion with PTI]{The quantitative results for PTI inversion are compared with e4e using different metrics.}
\vspace*{2mm}
\label{tab:tab_inv_pti}
\centering
\begin{tabular}{l c c c c}
Method & \multicolumn{1}{c}{FID $\downarrow$}  & \multicolumn{1}{c}{SSIM $\uparrow$} & \multicolumn{1}{c}{PSNR $\uparrow$}  &
\multicolumn{1}{c}{ACD $\downarrow$}\\
\midrule
e4e & 0.245 & 0.836 & 19.136 & 0.108 \\
PTI & 0.005 & 0.943 & 32.013 & 0.007  \\
\midrule
\end{tabular}
\end{table}
The~\ac{FID} scores decrease significantly from $0.245$ to $0.005$ compared to~\ac{e4e}, while~\ac{SSIM} scores improve by  $11.3\%$ from $0.836$ to $0.943$. The~\ac{PSNR} score also increases from $19.136$ to $32.013$, demonstrating a $40.2\%$ improvement. Finally, color composition scores,~\ac{ACD}, exhibit significantly enhanced performance, improving from $0.108$ to $0.007$.

\subsubsection{Latent Optimizer}
\begin{figure}[bp!]
  \begin{center}
    \includegraphics[width=80mm]{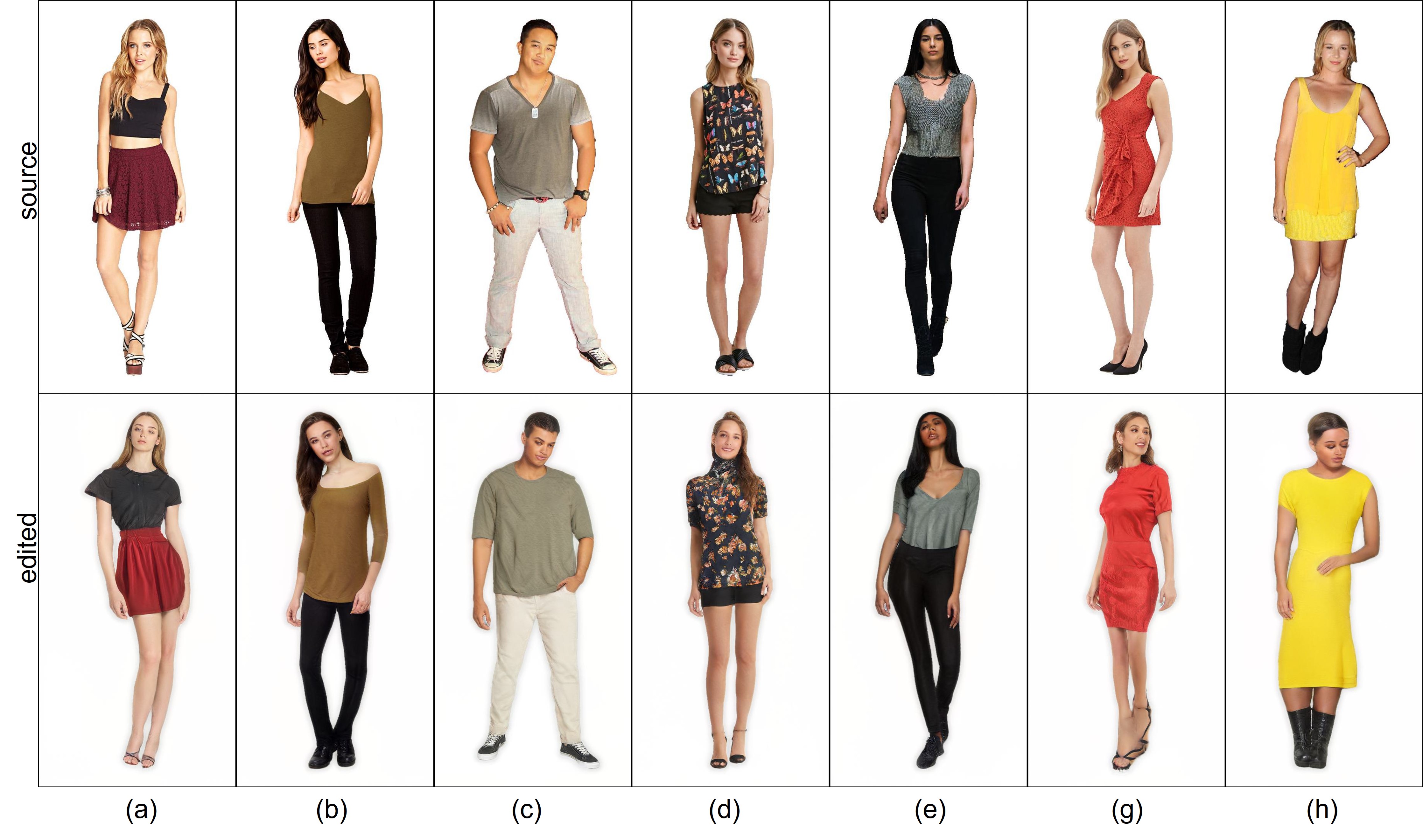}
  \end{center}
  \caption[Images for Latent optimizer scheme (Lengthen sleeves)]{The result for manipulating garments by lengthening sleeves with latent optimizer.}
  \label{fig:res_op}
\end{figure}

In this section, we present the outcomes of image manipulation using the latent optimizer approach. Figure~\ref{fig:res_op} displays a collection of sample images obtained by lengthening the sleeves of the garments using this approach. 
%==================================
%              Figure
%==================================
%==================================
The text prompt for this operation is “A long sleeve." The generated images demonstrate successful attribute modification, with the sleeves appearing longer in all cases. However, extensive distortion is observed in the results. The shape of the heads and faces are changed in all images of Figure~\ref{fig:res_op}(a-h), and the body configuration, including the posture of hands, feet, and main body, is not conserved. For example, in Figure~\ref{fig:res_op}(a), the angle of the hands and the posture of the legs are different from the original image. In Figure~\ref{fig:res_op}(g), the shape of the shoes deviates from the original ones, and the pattern of the clothing is not similar. Nonetheless, the color composition in all images is successfully maintained.
An extensive hyperparameter search is conducted to obtain the results, setting the $L_2$-norm term coefficient to one. The identity loss coefficient is deliberately chosen as a large value of $20$ to maintain the identity of the person in the edited images. However, the quantitative analysis shows a low preservation quality. Table \ref{tab:res_hair_sleeves} shows the quantitative comparison between different methods in the background region.  
%==================================
%              Table
%==================================
\begin{table}[bp!]
\renewcommand{\arraystretch}{1.0}
\setlength{\tabcolsep}{0.3em} 
\caption[Quantitative results for TD-GEM network (Lengthen sleeves)]{Comparison of different networks for lengthening sleeves in the background region.}
\vspace*{2mm}
\label{tab:res_hair_sleeves}
\centering
\begin{tabular}{l| c c c c c}
Methods &  FID $\downarrow$ &  SSIM $\uparrow$  & PSNR $\uparrow$  &
ACD $\downarrow$\\
\midrule
\multirow{1}{8em}{Latent Optimizer} 
  & 0.126 & 0.853 & 17.059 & 0.278  \\
\midrule
\multirow{1}{8em}{StyleCLIP Mapper} 
  & 0.021 & 0.919 & 24.293 & 0.165  \\
\midrule
\multirow{1}{4em}{TD-GEM} 
 & 0.030 & 0.935 & 27.543 & 0.146  \\
\midrule
\end{tabular}
\end{table}
A close examination of the table reveals that the FID score, when the latent optimizer is employed, is almost an order of magnitude higher in comparison to other methods. This suggests suboptimal results. Similarly, the SSIM and PSNR scores present a weaker performance. With respect to ACD, the latent optimizer also trails behind other methods. The ACD score for the latent optimizer stands at 68.5\% and 90.4\% higher than the scores recorded for StyleCLIP and TD-GEM, respectively, indicating a less desirable outcome. It is possible to improve the quality of the results of the latent optimizer by introducing more losses. However, the main disadvantage of the method is that an optimization problem must be solved for each individual image, which is pragmatically inconvenient. Therefore, we examine other methodologies that do not have this limitation in the next sections.

\subsubsection{StyleCLIP Mapper}
\begin{figure}[b]
  \begin{center}
    \includegraphics[width=80mm]{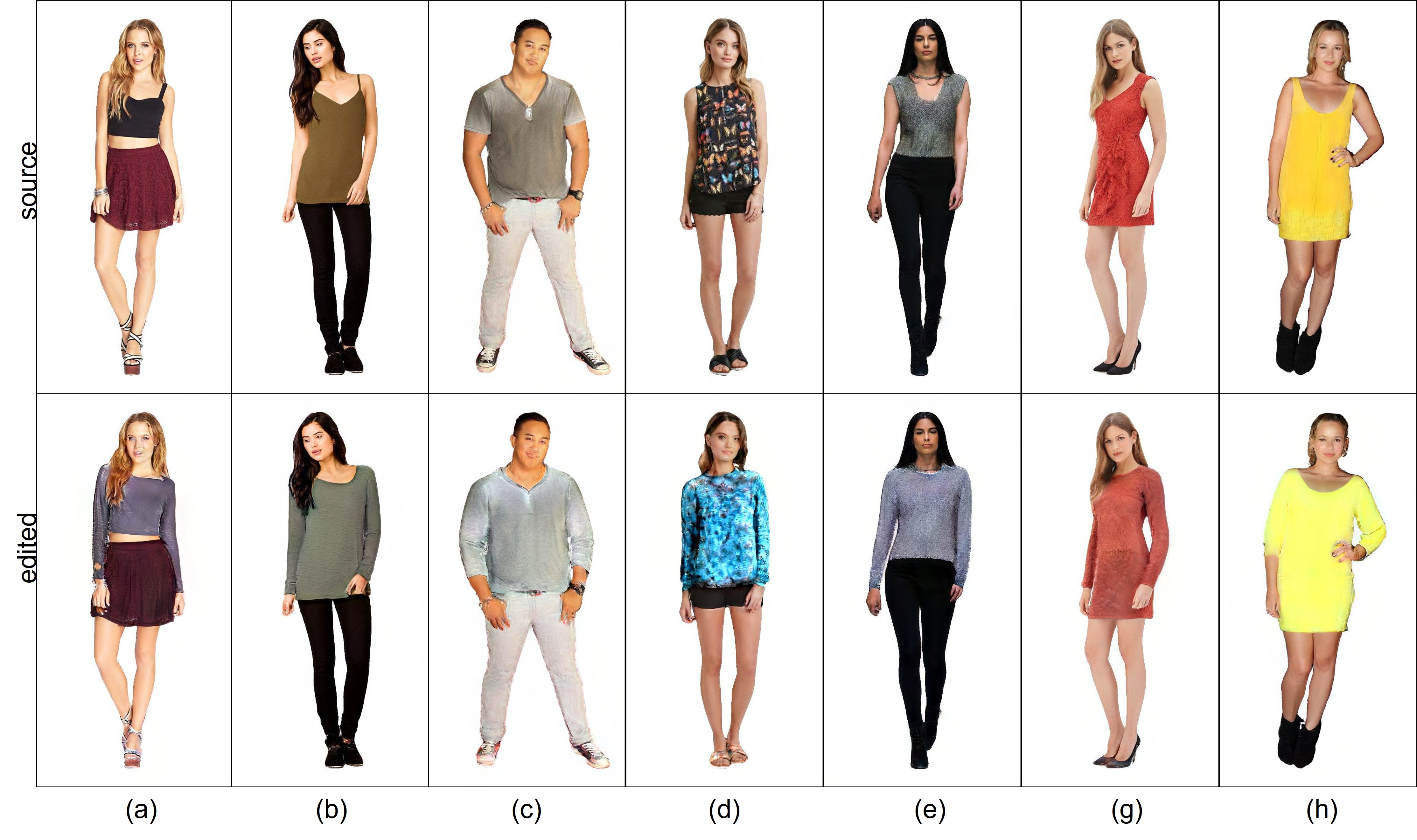}
  \end{center}
  \caption[Images for StyleCLIP mapper network (Lengthen sleeves)]{The StyleCLIP mapper network is utilized to extend the length of the sleeves.}
  \label{fig:res_mapper}
\end{figure}

The evaluation of the StyleCLIP mapper shows successful image manipulation, with the shape and configuration of hairs and faces maintained during editing (Figure \ref{fig:res_mapper}). However, there are some small deviations, such as the shape of the fingers in the left hand not being the same as in the source image (Figure \ref{fig:res_mapper}b). The color composition and pattern preservation are problematic, with the pattern of the clothing disappearing in some images and differences in color between the source and edited images being observed. The quantitative analysis reveals that the StyleCLIP mapper network has improved the background presentation (Table \ref{tab:res_hair_sleeves}), with better FID, SSIM, PSNR scores, and color composition compared to the latent optimizer scheme. However, a significant drawback of this methodology is that a separate mapper should be designed for each input description. This constraint is addressed in TD-GEM.
%==================================
%              Figure
%==================================
%==================================
\subsubsection{TD-GEM}
\begin{figure}[!b]
  \begin{center}
  \includegraphics[width=80mm]{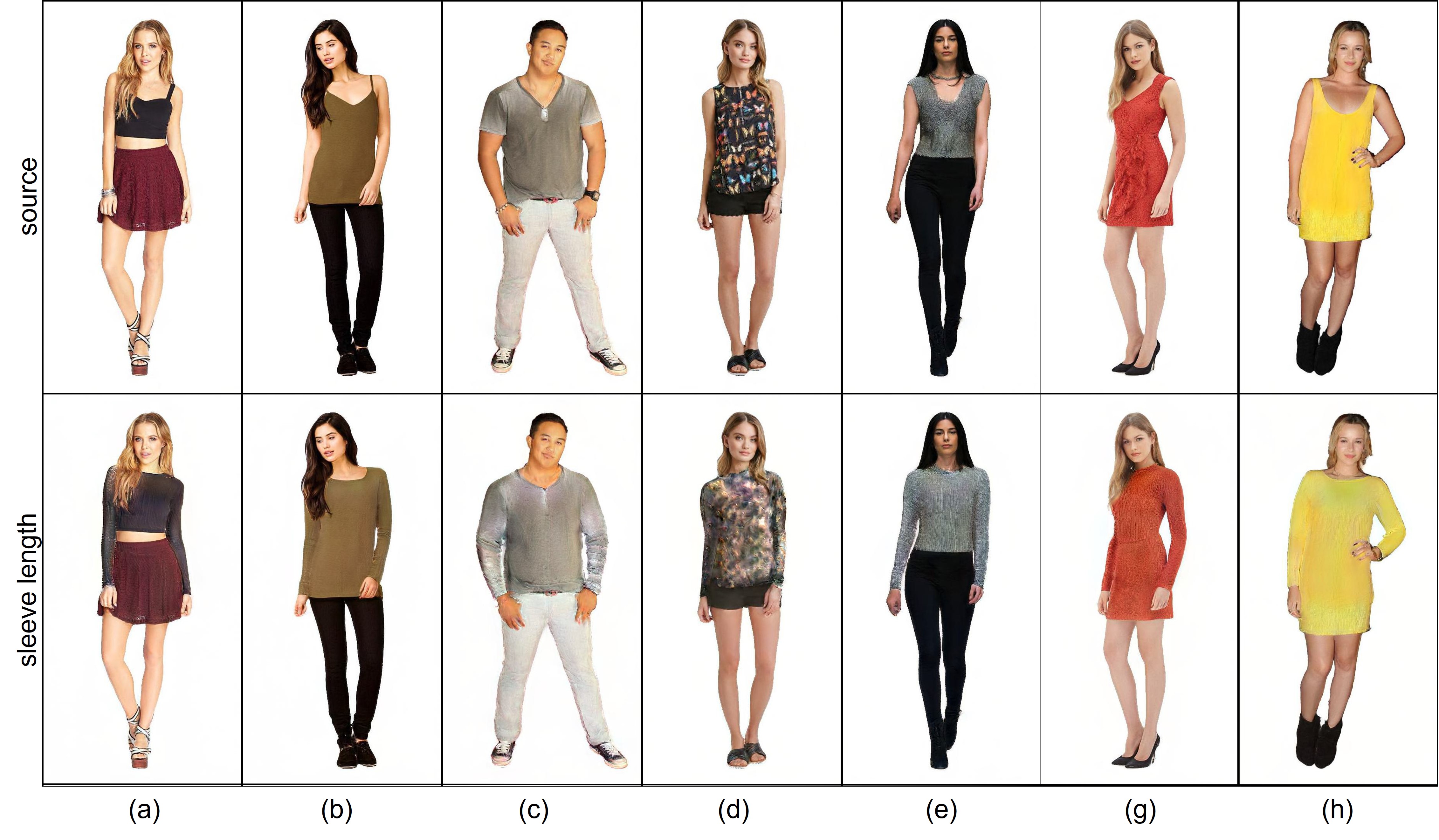}
  \end{center}
  \caption[Images for \ac{TD-GEM} network (Lengthen sleeves)]{The images are edited by~\ac{TD-GEM} network; the length of sleeves is increased.}
  \label{fig:res_hair_sleeves_set2}
\end{figure}

\begin{figure}[bp!]
  \begin{center}
    \includegraphics[width=80mm]{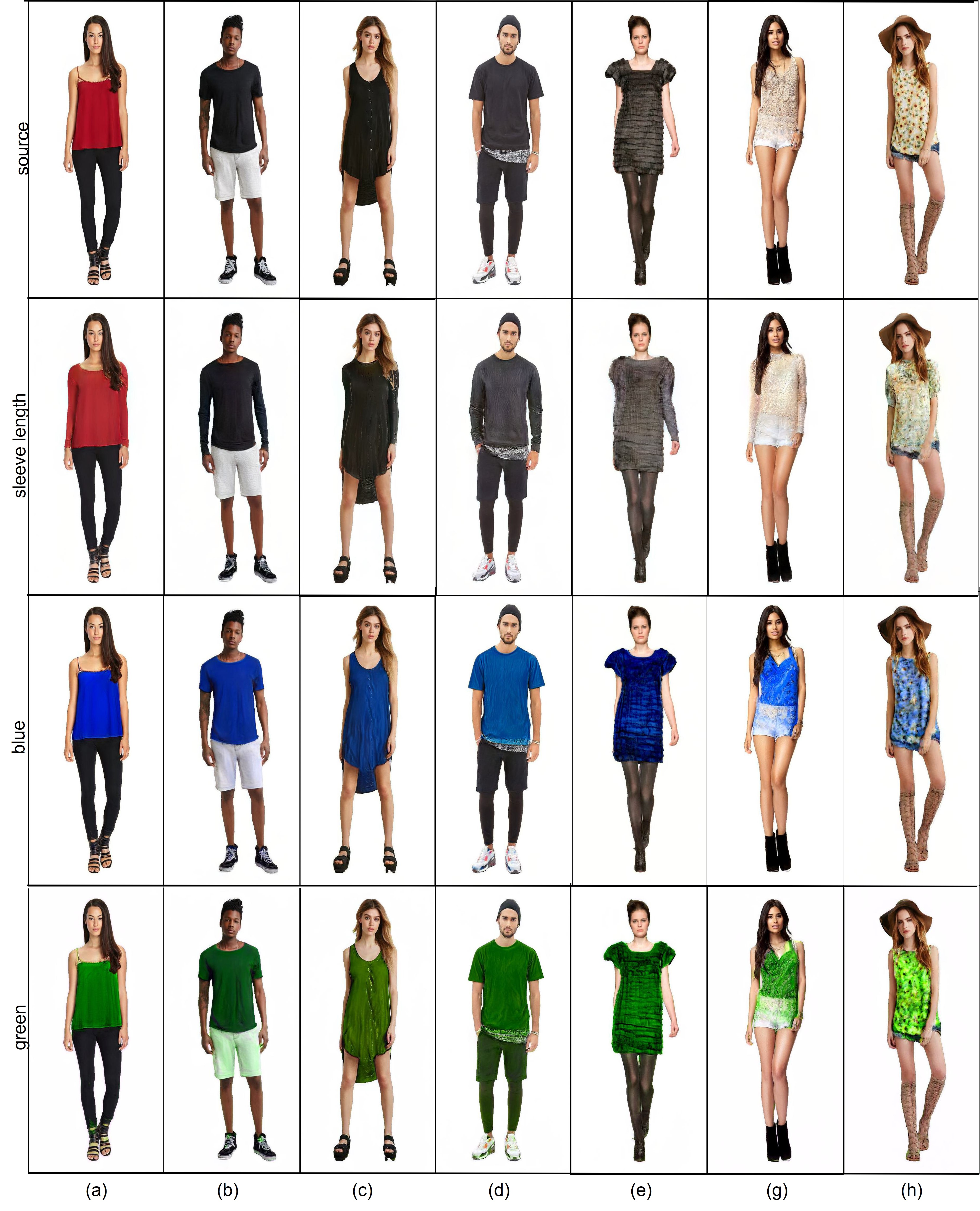}
  \end{center}
  \caption[Images for TD-GEM network (Blue color)]{The TD-GEM network is employed to alter the length of the sleeves and the color of the clothing to blue and green.}
  \label{fig:res_hairclipset2}
\end{figure}

This section presents the results of \ac{TD-GEM} network for manipulation of full-body human images in the fashion domain. Figure~\ref{fig:res_hair_sleeves_set2} shows a set of qualitative results from this method.
%==================================
%              Figure set1 and set2
%==================================
%==================================
The degree of disentanglement in the generated images is quite acceptable, with preserved details in the person's configuration, including hands, legs, and main body, which have the same postures, shapes, and contours as the original images. In contrast to the previous StyleCLIP mapper network, the finger details are well preserved in the edited images, as seen in Figure~\ref{fig:res_hair_sleeves_set2}b. Additionally, the faces and hair in the manipulated images are indistinguishable from the original images due to interpolation between the original and edited images, which improves the quality of those regions. The color composition in the TD-GEM network is remarkably better than the previous StyleCLIP mapper network, as seen in Figure~\ref{fig:res_hair_sleeves_set2}h, where only slight deviations between the original yellow and edited yellow are observed.
Furthermore, to assess the preservation of the shape and patterns of the clothing, several samples are presented in  Figure~\ref{fig:res_hairclipset2}.
%==================================
%              Figure blue
%==================================
The results indicate that the pattern of the garment is well-preserved, even for complicated patterns with more detailed garments, as seen in Figure~\ref{fig:res_hairclipset2}(d-e), where the clothing has a more complicated form and wrinkles.
A quantitative comparison between different approaches, including latent optimizer, StyleCLIP mapper, and \ac{TD-GEM}, is presented in Table~\ref{tab:res_hair_sleeves}. The scores on the background area represent the degree of disentanglement in the image manipulation. In the~\ac{TD-GEM} case, the background is preserved with the same quality as the StyleCLIP mapper case, with an FID score of $O(10^{-2})$ for both cases. Although the FID score in \ac{TD-GEM} is worse than the StyleCLIP mapper case, the SSIM and PSNR scores show an improvement. The color composition in both cases is almost the same, where the better results belong to the~\ac{TD-GEM} network. 

In this paper, the TD-GEM network is utilized for the color manipulation of garments. The last two experiments are conducted where the color is changed to blue and green, respectively (last two rows). The body configurations, faces, and hairs are well preserved, while the garment color is changed to blue. The degree of color saturation varies, but the color attribute is successfully modified in all images. Color leakage is observed in one image, where the trousers are painted green (Figure~\ref{fig:res_hairclipset2}b). The results indicate that the patterns of garments are well-preserved, even for complicated patterns with more detailed garments, as seen in Figure~\ref{fig:res_hairclipset2}(d-h), where the garment has a more complicated shape and wrinkles.

\begin{figure}[bp!]
  \begin{center}
    \includegraphics[width=80mm]{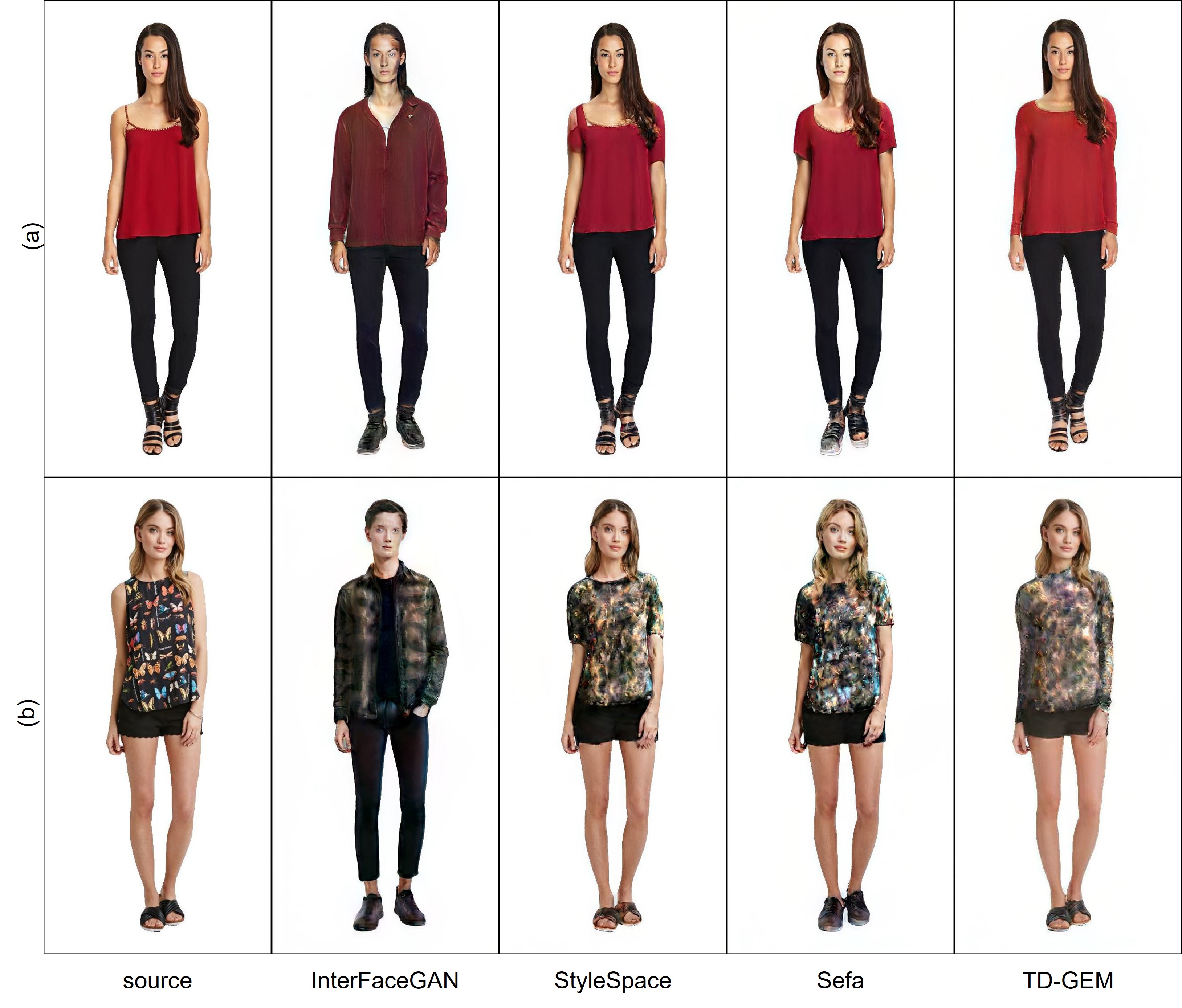}
  \end{center}
  \caption[Comparison of TD-GEM with recent methodologies]{The figure compares various methodologies for manipulating human clothing, including InterFaceGAN, StyleSpace, Sefa, and the TD-GEM network. The first three methods are based on the work by~\cite{fu2022stylegan}. Two samples from the testing dataset are shown in the two rows (a) and (b).}
  \label{fig:res_stylehuman}
\end{figure}

The quantitative comparison of color manipulation using \ac{TD-GEM} is provided in Table~\ref{tab:res_hair_sleeves_blue_green}. The scores for the background region are almost the same for both colors. The SSIM and PSNR values are also very close. However, a higher value of the color score for the green color in the background could be related to the color leakage effect. 
%It is worth noting that the background scores indicate the level of disentanglement in image manipulation. Additionally, the assessment of foreground image manipulation relies on qualitative results, as the metrics inadequately capture the quality of editing in the foreground.
%==================================
%              Table
%==================================
\begin{table}[bp!]
\renewcommand{\arraystretch}{1.0}
\setlength{\tabcolsep}{0.3em} 
\caption[Quantitative results for TD-GEM network (Color and sleeves)]{Result for TD-GEM network by changing the color in the background region} 
\vspace*{2mm}
\label{tab:res_hair_sleeves_blue_green}
\centering
\begin{tabular}{l| c c c c c}
Text &  FID $\downarrow$ &  SSIM $\uparrow$  & PSNR $\uparrow$  &
ACD $\downarrow$\\
\midrule
\multirow{1}{4em}{sleeve}
 & 0.030 & 0.935 & 27.543 & 0.146  \\
\midrule
\multirow{1}{4em}{blue } 
 & 0.017 & 0.956 & 31.607 & 0.191  \\
\midrule
\multirow{1}{4em}{green}
 & 0.030 & 0.956 & 31.608 & 0.294  \\
\midrule
\end{tabular}
\end{table}
A comparison between the \ac{TD-GEM} and supervised and unsupervised editing methods by~\cite{fu2022stylegan} is shown in Figure~\ref{fig:res_stylehuman}.
%==================================
%              Figure styleganhuman
%================================
%===============================
Two samples from the test dataset are edited using four approaches, and the results are illustrated in the top and bottom rows. InterFaceGAN radically changes the shape and form of the garment, while StyleSpace provides a better solution but still has some issues with the form of the clothing. Sefa presents promising results, but the disentanglement property is not successfully enforced. The shape of the shoes is not preserved, as can be seen in Figure~\ref{fig:res_stylehuman}b. On the other hand, the \ac{TD-GEM} performs the best by successfully lengthening sleeves up to the wrist and keeping the other unrelated attributes untouched.
The quantitative scores for TD-GEM are compared with the three aforementioned methods in Table~\ref{tab:res_hair_sleeves_stylehuman}. FID scores show an order of magnitude better performance for~\ac{TD-GEM} in the background region. The SSIM and PSNR scores are also substantially higher for this approach. ACD scores are superior to InterFaceGAN and Sefa, and similar to StyleSpace. 
%==================================
%              Table
%==================================
\begin{table}[bp!]
\renewcommand{\arraystretch}{1.0}
\setlength{\tabcolsep}{0.3em} 
\caption[Quantitative results for InteFaceGAN, StyleSpace and Sefa]{The quantitative scores for the comparison of TD-GEM with IntefaceGAN, StyleSpace, and Sefa through lengthening sleeves in the background region.} 
\vspace*{2mm}
\label{tab:res_hair_sleeves_stylehuman}
\centering
\begin{tabular}{l| c  c c c c}
Method   &  FID $\downarrow$ &  SSIM $\uparrow$  & PSNR $\uparrow$  &
ACD $\downarrow$\\
\midrule
\multirow{1}{4em}{TD-GEM} 
  & 0.030 & 0.935 & 27.543 & 0.146  \\
\midrule
\multirow{1}{6em}{InterFaceGAN }
  &  0.199& 0.864 & 16.794 & 0.712   \\
\midrule
\multirow{1}{6em}{Style Space} 
  & 0.102 & 0.898 & 22.725 & 0.137 \\
\midrule
\multirow{1}{4em}{Sefa} 
  & 0.176 & 0.882 &  20.540 & 0.244  \\
\midrule
\end{tabular}
\end{table}
%==================================
It's important to highlight that our measurement scores may not provide a precise assessment of the quality in the foreground; hence the comparison was conducted using the outcomes of qualitative analysis. As per our analysis, the approach we proposed yields superior outcomes when compared to the baseline methodologies.
\subsection{Ablation Study}
This section investigates the efficacy of two key assumptions: identity loss and semantic injection across all layers. 

\begin{figure*}[b]
  \begin{center}
    \includegraphics[height=8 cm]{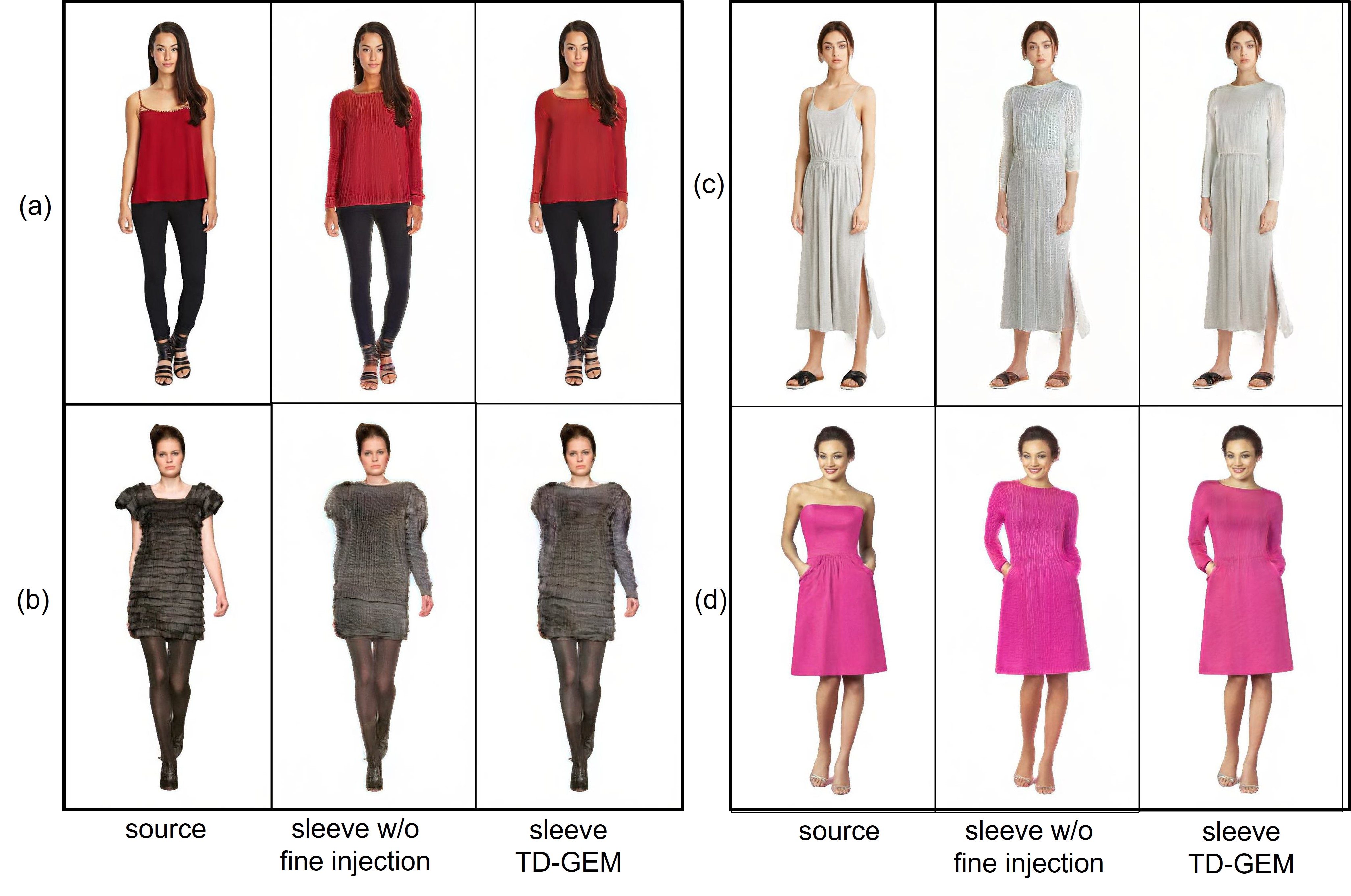}
  \end{center}
  \vspace*{-3mm}
  \caption[I)]{The effect of semantic injection into the fine module in  lengthening the sleeves.}
  \label{fig:abl_fine_inj}
\end{figure*}

We first analyze the impact of incorporating semantic injection into all three modules while editing the sleeve length, as demonstrated in Figure \ref{fig:abl_fine_inj}. 
%===============================
%              Figure
%===============================
%================================
In case (a) (top row), the image generated without fine injection exhibit artificial vertical straw compared to the ground truth. However, this artifact disappears when all three modules are employed. In case (b), the presence of all three modules leads to better preservation of the wrinkle. Table \ref{tab:abl_no_injection} shows quantitatively the benefits of applying the semantic injection into the fine module. 
\begin{table}[bp!]
\renewcommand{\arraystretch}{1.0}
\setlength{\tabcolsep}{0.3em}
\caption[O]{The quantitative scores for the case with and without injection in the fine module for the lengthening sleeves in the background region.} 
\vspace*{3mm}
\label{tab:abl_no_injection}
\centering
\begin{tabular}{l| c  c c c c}
Method &  FID $\downarrow$ &  SSIM $\uparrow$  & PSNR $\uparrow$  &
ACD $\downarrow$\\
\midrule
TD-GEM 
 & \textbf{0.030} & \textbf{0.935} & \textbf{27.543} & \textbf{0.146}  \\
\midrule
w/o fine injection 
  &0.089& 0.925 & 26.883 & 0.209   
\end{tabular}
\end{table}
It highlights that this step helps to better keep the background details in the edited image, as all metrics are improved.

\begin{figure}[htb!]
  \begin{center}
    \includegraphics[width=80mm]{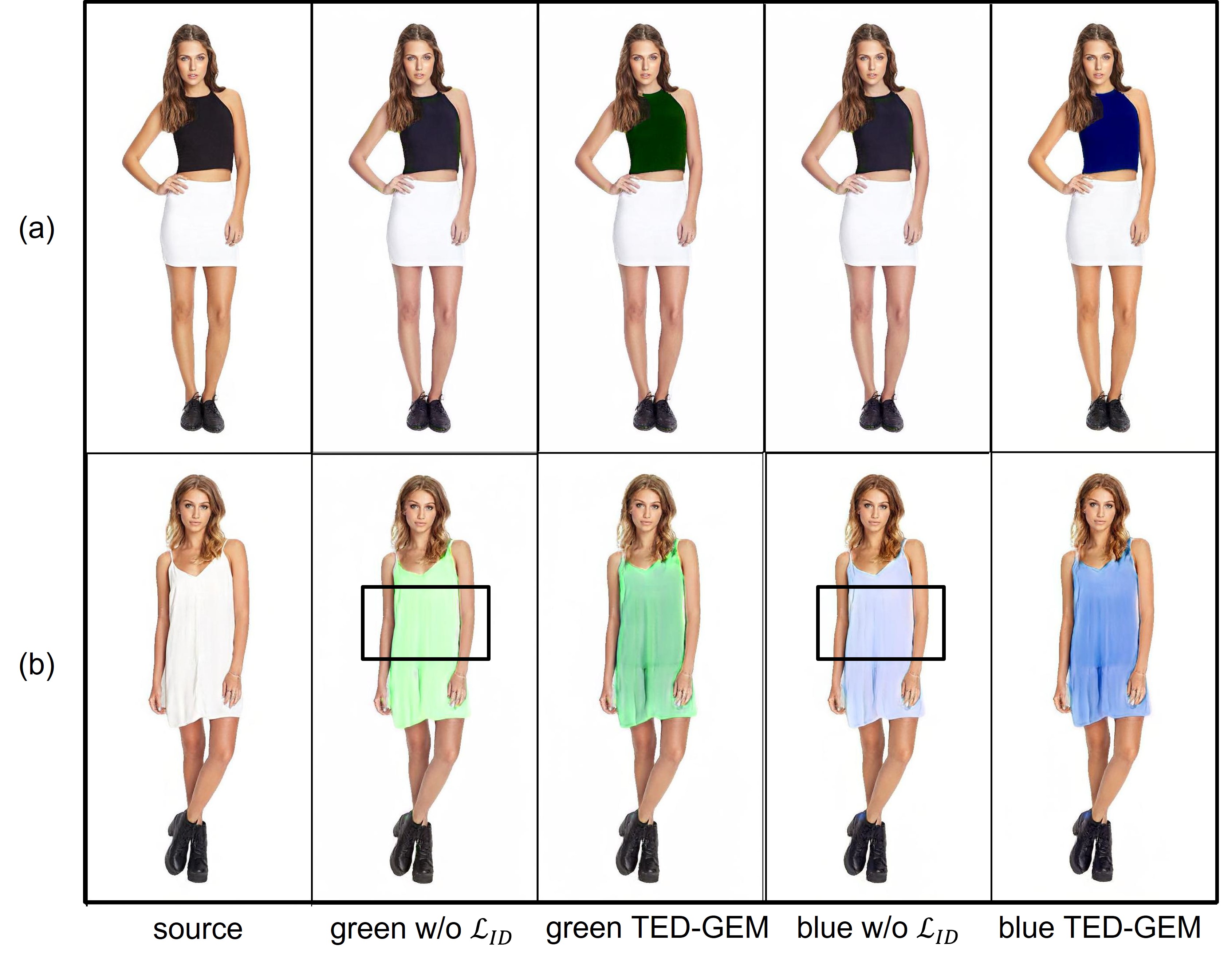}
  \end{center}
  \caption[I)]{The effect of identity loss in editing the color.}
  \label{fig:abl_no_id}
\end{figure}

Next, we conduct a second ablation study focusing on the role of identity loss. Figure \ref{fig:abl_no_id} displays edited images both with and without the identity term, taking into account the input description that specifies color alterations to either blue or green.
%==================================
%              Figure
%==================================
%================================
When the identity loss is omitted, the color change appears muted in both cases (a) and (b). However, when we integrate identity loss, it shows enhanced preservation of the original garment's characteristics. In the case of lengthening the sleeve, the effect of including identity loss appears to be trivial in our experiments.

\section{Discussion and Conclusion}
This paper has presented a novel approach for full-body human fashion image editing through textual input descriptions. The image manipulation process involves two stages: firstly, obtaining a latent representation of the image in the latent space of a pre-trained StyleGAN2-ADA network, and secondly, editing the image by navigating semantically along with the relevant directions using a pre-trained language model CLIP. To this end, two~\ac{GAN} inversion methods, e4e and PTI, have been utilized to acquire the latent representation, and it has been found that PTI produced superior accuracy, albeit at a higher computational cost. To proceed with attribute editing, our proposed~\ac{TD-GEM} manipulates a sleeve length and color of a garment, integrating new loss functions that accomplish full-body human image editing. We discovered that incorporating semantic injection into the fine module enhances image editing outcomes, while the impact of identity loss remains relatively insignificant. Extensive experiments demonstrate that our solution can edit high-quality fashion images. It outperforms competing methods, such as latent optimizer and StyleCLIP mapper network, on evaluated metrics in terms of accuracy and reducing computational complexity due to training only one network for color or sleeve-length textual descriptions. Furthermore, the findings of our study will serve as a benchmark for future processes involving full-body human fashion image editing and can be expanded upon with varying textual prompts such as “shortening sleeve length", “adding stripes", and “incorporating patterns" among others.

\bibliographystyle{icml2021}

%\appendix
%\section{Do \emph{not} have an appendix here}

%\textbf{\emph{Do not put content after the references.}}
%

%%%%%%%%%%%%%%%%%%%%%%%%%%%%%%%%%%%%%%%%%%%%%%%%%%%%%%%%%%%%%%%%%%%%%%%%%%%%%%%
%%%%%%%%%%%%%%%%%%%%%%%%%%%%%%%%%%%%%%%%%%%%%%%%%%%%%%%%%%%%%%%%%%%%%%%%%%%%%%%

\end{document}